\RequirePackage{fix-cm}
\documentclass[smallextended]{svjour3}       

\smartqed  
\usepackage{graphicx}
\usepackage{subfig}
\usepackage{todonotes}
\usepackage{multirow}
\usepackage{natbib}
\usepackage{url}
\begin{document}

\title{Developing a Fidelity Evaluation Approach for  Interpretable Machine Learning
}

\titlerunning{Developing a Fidelity Evaluation Approach}        

\author{Mythreyi Velmurugan \and Chun Ouyang \and Catarina Moreira \and Renuka Sindhgatta}


\institute{M. Velmurugan, C. Ouyang, C. Moreira, R. Sindhgatta \at
              Queensland University of Technology, Brisbane, Australia \\
              \email{\{m.velmurugan,c.ouyang,catarina.pintomoreira,renuka.sr\}@qut.edu.au}           
}

\date{Received: date / Accepted: date}

\maketitle

\begin{abstract}
Although modern machine learning and deep learning methods allow for complex and in-depth data analytics, the predictive models generated by these methods are often highly complex, and lack transparency. Explainable AI (XAI) methods are used to improve the interpretability of these complex models, and in doing so improve transparency. However, the inherent fitness of these explainable methods can be hard to evaluate. In particular, methods to evaluate the fidelity of the explanation to the underlying black box require further development, especially for tabular data. In this paper, we (a) propose a three phase approach to developing an evaluation method; (b) adapt an existing evaluation method primarily for image and text data to evaluate models trained on tabular data; and (c) evaluate two popular explainable methods using this evaluation method.  Our evaluations suggest that the internal mechanism of the underlying predictive model, the internal mechanism of the explainable method used and model and data complexity all affect explanation fidelity. Given that explanation fidelity is so sensitive to context and tools and data used, we could not clearly identify any specific explainable method as being superior to another.
\end{abstract}

\section{Introduction}
\label{sec:intro}
Modern machine learning and deep learning techniques have allowed for analytics of complex data and enable decision-making~\citep{Moreira20medical}.
However, these advanced machine learning techniques often hamper human interpretability, therefore lacking transparency, affecting the safety of and fairness towards stakeholders. Explainable AI (XAI) methods are used in order to improve the interpretability of these complex ``black box'' models, thereby increasing transparency and enabling informed decision-making~\citep{Guidotti2018}.

Despite this, methods to assess the quality of explanations generated by such explainable methods are so far under-explored. In particular, functionally-grounded evaluation methods, which measure the inherent ability of explainable methods in a given situation, are often specific to a particular type of dataset or explainable method. A key measure of functionally-grounded explanation fitness is explanation fidelity, which assesses the correctness and completeness of the explanation with respect to the underlying black box predictive model~\citep{Zhou2021}.

Evaluations of fidelity in literature can generally be classified as one of the following: external fidelity evaluation, which assesses how well the prediction of the underlying model and the explanation agree, and internal fidelity, which assesses how well the explanation matches the decision-making processes of the underlying model~\citep{Messalas2019}. While methods to evaluate external fidelity are relatively common in literature ~\citep{Guidotti2019a,Lakkaraju2016,Ming2019,Shankaranarayana2019}, evaluation methods to evaluate internal fidelity using black box models are generally limited to text and image data, rather than tabular~\citep{Du2019,Fong2017,Nguyen2018,Samek2017}.


In this paper, we propose a novel evaluation method based on a three phase approach: (1) the creation of a fully transparent, inherently interpretable white box model, and evaluation of explanations against this model; (2) the usage of the white box as a proxy to refine and improve the evaluation of explanations generated by a black box model; and (3) test the fidelity of explanations for a black box model using the refined method from the second phase.

The main contributions of this work are as follows:
\begin{enumerate}
    \item The design of a three phase approach to developing a fidelity evaluation method. 
    \item The development of a fidelity evaluation method for black box models trained on tabular data.
    \item The evaluation of two common and popular local feature attribution explainable methods (LIME and SHAP) using the proposed evaluation approach and method.
\end{enumerate}

This paper is structured as follows. The following section (section~\ref{sec:background}) will provide a more detailed introduction to the field of XAI, as well as evaluations of explanations. Section~\ref{sec:approach} will outline the fidelity evaluation undertaken as part of this work, and section~\ref{sec:setup} will describe the experimental setup used to implement this approach. The results of the experiments are outlined in section~\ref{sec:results} and section~\ref{sec:conclusion} will conclude this work.









\section{Background and Related Works}
\label{sec:background}
\subsection{Explainable AI}
The ``black box problem'' of AI arises from the inherent complexity and sophisticated internal data representations of many modern machine learning and AI algorithms. While prediction models created by such algorithms may produce more accurate results, this accuracy comes at the cost of human interpretability of these prediction models~\citep{Koska2019}. As such, the research theme of XAI has arisen in order to provide the ability to interpret the decision making of black box prediction models, in order to ensure system quality and facilitate informed decision-making~\citep{Carvalho2019}.

In this paper, the terms ``interpretability'' and ``explanations'' will be used as defined by~\citet{Guidotti2018}, where ``interpretability'' refers to the ability to provide meaning in human understandable terms, and ``explanations'' are the interface between the human and the predictive model. Interpretability in machine learning is generally broken down into two categories: interpretable prediction models and post-hoc interpretation. Interpretable prediction models are those that are generated in such a way as to be immediately interpretable by a human~\citep{Guidotti2018}, though this often means that the models are simpler, and so may have reduced predictive power. 
Examples of inhrently interpretable models include decision trees and linear regression models, where either the decision path or the weights of the model can be returned to the user, allowing them to understand the inner workings of the model.
Post-hoc methods provide interpretations that are not inherent to the prediction model, but are applied after its creation. 
Post hoc methods are usually applied to \textit{opaque} models, which correspond to models that are not interpretable by design, and whose internal mechanisms are complex enough to hamper human interpretability or are simply not visible. Examples of such models include neural networks or tree-ensemble models, such as random forest models.
Although the use of post-hoc interpretability allows for the creation of more complex, and accurate, predictive models without compromising the interpretability of the model, given that post-hoc interpretation mechanisms are external to the prediction model, there is no guarantee that the provided explanations are fully and correctly representative of the prediction model's decision making, or otherwise fit for purpose~\citep{Rudin2018}.  

In post-hoc XAI, most interpretable algorithms focus on feature attribution for local explanations.
A local explanation, or outcome explanation, is an explanation provided not for the behaviour of the entire predictive model, but explanation for a single model prediction, or the model's behaviour within a small neighbourhood of data surrounding a particular input~\citep{Guidotti2018}. Feature attribution explanations, or predict parts explanation tasks, are explanations that present the impact of each input feature on the end result~\citep{Maksymiuk2020}.

Typical examples of this class of explainable methods are \emph{Local Interpretable Model-Agnostic Explanations (LIME)}~\citep{Ribeiro2016} and \emph{SHapley Additive exPlanation (SHAP)}~\citep{Lundberg2017}, though they use different mechanisms to interpret predictive models. LIME creates surrogate models to mimic part of the decision-making of the underlying predictive model, and use this surrogate model to extract explanations about the predictive model's decision-making. In order to do so, LIME first creates new input instances by randomly sampling instances within the neighbourhood of the original instance, and creates predictions for these from the predictive model. A surrogate model is then trained using these predictions as the ground truth. A known downfall of this approach is that, as the number of features in the input increases, so does the number of possible instances within the neighbourhood. Given that only a limited number are sampled to train the surrogate model, this often results in different instances being sampled every time an explanation is generated, leading to unstable explanations~\citep{Visani2020}. LIME is fully model-agnostic and is not restricted to any class of predictive model. On the other hand, SHAP takes a game-theoretic approach to model explanation. Using the coalitional game theory concept of Shapley Values, SHAP attempts to determine the marginal contribution of each feature to the final output of the underlying predictive model. 

\subsection{Evaluating Explanation Fidelity}
In this work, we focus on evaluating post-hoc explainability approaches. In particular, the evaluation method we aim to develop will attempt to assess local, feature attribution explanations.
In XAI, evaluation methods are categorised in three levels of evaluation proposed by~\citet{DoshiVelez2017}:
\begin{itemize}
    \item \emph{Application-grounded evaluation:} wherein the evaluation is conducted with real end users in full context replicating a real-world application;
    \item \emph{Human-grounded evaluation:} wherein the evaluation is conducted with laymen in simpler simulated contexts or completing proxy tasks that reflect a target application context; and
    \item \emph{Functionally-grounded evaluation:} which requires no input from users and relies on evaluating the inherent abilities of the system via a formal definition of interpretability.
\end{itemize}

The evaluation method developed in this work focuses on this latter category of functionally-grounded evaluation. Specifically, we will develop an evaluation method to assess \textit{explanation fidelity} with respect to the original predictive model. The evaluation of fidelity in many ways remains an open question in the field of XAI. The terms ``fidelity" is generally used to refer to a measure of how faithful and relevant an explanation is to the underlying black box model~\citep{Yang2019}. As such, fidelity is an integral requirement of explanations, given that it is a measure of explanation accuracy, and lack of fidelity in explanations could hamper decision-making, or enable misinformed decision-making. \citet{Markus2021} and \citet{Zhou2021} divide fidelity into two characteristics of explanations that can be measured: the \textit{completeness} of an explanation in capturing the dynamics of the underlying model; and the correctness and truthfulness of the explanation with respect to the underlying model (\textit{soundness}).

Although a number of methods have been proposed to assess fidelity, they are often highly specific to particular methods or datasets. \citet{Messalas2019} breaks down fidelity evaluation approaches into two general characteristics: measurements of \textit{external fidelity} and \textit{internal fidelity}. They refer to external fidelity approaches as those that compare how often the decision implied by the explanation and the decision made by the underlying predictive model agree. For example, when evaluating surrogate models that approximate the underlying predictive model or explanation methods that use surrogate models, this method compares the accuracy of the surrogate model's predictions using the predictions of the underlying model as ground truth, as done by ~\citet{Shankaranarayana2019}. Explainable methods that produce decision rules also often evaluated through this approach, where the predictions of the decision rule or decision rule sets are compared against the black box model's predictions for the relevant instances~\citep{Guidotti2019a,Lakkaraju2016,Ming2019}. While this approach can be used to measure the fidelity of the explanation's predictions with respect to the predictions of the underlying model, this does not guarantee that the explanations are faithful of the decision-making process of the black box model (for example, importance given to each feature). Moreover, some explainable methods do not produce some output that can be used for external fidelity evaluation approaches, such as a surrogate model or a set of decision-rules.

As such, we aim to develop a method to assess the internal fidelity of feature attribution explainable methods, where the fidelity of the explanation regarding the underlying model's decision-making process is evaluated~\citep{Messalas2019}. Given that the internal workings of a black box model are opaque, significant effort is required to assess the internal fidelity of an explanation or explainable method. Potential internal fidelity evaluation approaches include:
\begin{enumerate}
    \item Generating explanations for a fully transparent, inherently interpretable ``white box'' model, and comparing the explanation to the white box model's decision-making~\citep{Ribeiro2016}.
    \item Using a post-hoc explanation approach to explain both a surrogate model and the underlying predictive model, and comparing how often the explanations concur~\citep{Messalas2019}.
    \item ``Removing" or perturbing the features identified as relevant by the explanation and measuring the change in model output~\citep{Du2019,Fong2017,Nguyen2018,Samek2017}. 
\end{enumerate}

None of these three methods is currently suitable to evaluate feature attribution explanations of tabular data. The second approach is intended for the evaluation of custom, stand-alone surrogate models used to approximate the whole dynamic of predictive models, and so, is not suitable for evaluating existing, local post-hoc approaches, as we aim to do. While the first approach can be useful for simple comparison of different explainable methods or approaches, as was done by~\citet{Ribeiro2016}, high fidelity for a relatively simple white box model does not imply high fidelity for a more complex and sophisticated black box model. And though the third approach appears to be the most promising, it is typically currently only applied to evaluate explanations of models trained on text or image data. Alteration is relatively simple in these cases, where the relevant pixels can be replaced with a constant value, removed, blurred or randomised in image data~\citep{Fong2017,Samek2017}, or relevant sentences, phrases or words can be removed from the input in text data~\citep{Du2019,Nguyen2018}. However, this approach becomes more complex for tabular data, where removal of features simply results in the predictive model substituting in an alternative value. While we have previously proposed an adaptation of this approach for tabular data~\citep{Velmurugan2021}, more extensive adaptations and validation of the adapted method are needed. For this purpose, we combine the first and third approaches into three phases, as described in section~\ref{sec:approach}.

\section{Fidelity Evaluation Approach}\label{sec:approach}
We used a three-phase approach to develop an appropriate method to evaluate the fidelity of a black box trained on tabular data. This approach begins with the creation of a fully transparent, inherently interpretable white box model (in this case, a Decision Tree), and the evaluation of explanations directly against this white box model. Following this, the white box is used as a proxy for a black box model to develop and refine a method to evaluate explanations for a black box model, which ensures that the method is suitable and accurate. In the final phase, this developed method is applied to test the fidelity of explanations for a black box model.






\subsection{Phase 1: Determining Fidelity using White Box Model}\label{sec:phase-1}

The initial phase attempts to determine the fidelity of the explainable method in interpreting an inherently interpretable, fully transparent white box model, where the decision making of the predictive model is fully clear, such as a simple linear model or decision tree. The method used for this phase was adapted from the method used by~\citet{Ribeiro2016} in their evaluations of LIME. As the original method was applied to text data, a number of adaptations are required to make this method more suitable for tabular data. 

The basic approach of this method is to, for any given instance, compare the most relevant features as determined by the predictive model (the ``true features'') to the most relevant features as determined by the explanation (the ``explanation features''). The predictive model is first trained. In the original method,~\citet{Ribeiro2016} restricted the number of features that the model relied on when making its prediction.  However, given that different tabular datasets represent different problems with varying numbers of features, we do not think that restricting the number of features to a fixed amount would be a feasible approach for the proposed evaluation.

Two metrics can be applied to evaluate fidelity. Firstly, \citet{Ribeiro2016} used recall ($\mathcal{R}$) as a measure of fidelity for this method, which is defined as follows, where the term \textit{True Features} represents the relevant features as extracted directly from the white box model and \textit{Explanation Features} represents the features characterised as most relevant by the explanation:

\begin{equation}\label{eq:recall}
    \mathcal{R} = \frac{|True~Features \cap Explanation~Features|}{|True~Features|}
\end{equation}

This measure is an indication of how well the explanation captures the most relevant features from the predictive model, i.e. as a measure of completeness of the explanation. Additionally, to understand how well the explanation excludes irrelevant features (soundness of the explanation), precision ($\mathcal{P}$) can be measured:

\begin{equation}\label{eq:precision}
    \mathcal{P} = \frac{|True~Features \cap Explanation~Features|}{|Explanation~Features|}
\end{equation}

Both metrics are applied at the instance level, and can be averaged over the dataset.

\begin{figure}[h!]
\resizebox{\columnwidth}{!} {
    \centering
    \includegraphics{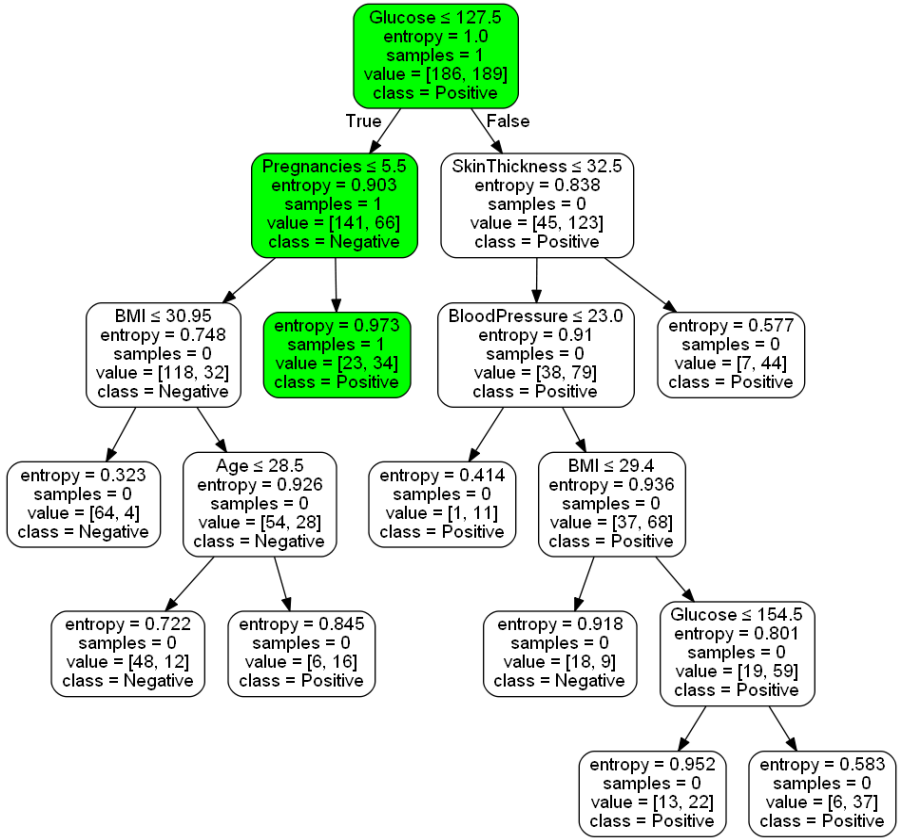}
    }
    \caption{Prediction of Diabetes, using a decision tree. The highlighted decision path shows all the "True Features" that are impacting the prediction for a single instance.}
    \label{fig:tree_path}
\end{figure}

The extraction of the true features and explanation features differs depending on the predictive model and explainable methods used. For a standard, single decision tree predictive model, the most appropriate approach to extract the true features for any given instance is to identify the set of unique features that fall along the decision path for that instance. For example, for the tree in Figure~\ref{fig:tree_path}, the model has returned a result of ``Positive" (diabetic) based on an individual's test results because their glucose level is under 127.5 and they have experienced more than 5 pregnancies. Thus, the "true features" for this instance are "Glucose" and "Pregnancies". For feature attribution explanations, or predictive models that apply a coefficient to each feature, the features can be sorted by this weighting and the most appropriate number of features can be chosen. To test recall of the most relevant features when using a decision tree, an appropriate sample of explanation features is required. 

This sampling may be different for each metric being used, particularly when extracting explanation features. For example, to ensure that the sample size of features is appropriate when measuring recall, a reasonably large set of explanation features is required. If the underlying predictive model is a decision tree, the length of the longest decision path in the tree can be used as the appropriate sample size for explanation features. Alternatively, or when this length is larger than the feature space on the decision tree, two-thirds of the feature space can be extracted. To test precision, only the most relevant features are necessary. Therefore, the most appropriate features to extract are those that fall into the top quartile, based on the feature weights produced as part of the explanation. That is, this set of Q1 features does not refer to a number of features equivalent to 25\% of the feature space, but to all features with weights that fall into the top quartile of the feature weight distribution within the explanation. 

Once both the set of true features and the set of explanation features have been extracted for each instance, the appropriate metric can be applied for those sets. Given the differing sizes of the explanation feature sets used for each metric, the F1 score cannot be calculated as a measure of fidelity for each instance. Thus, there are two metrics of fidelity for each instance: recall to calculate the completeness of the explanation; and precision to determine the correctness of the explanation.

\begin{figure}[h!]
    \resizebox{\columnwidth}{!} {
    \includegraphics[scale=0.9]{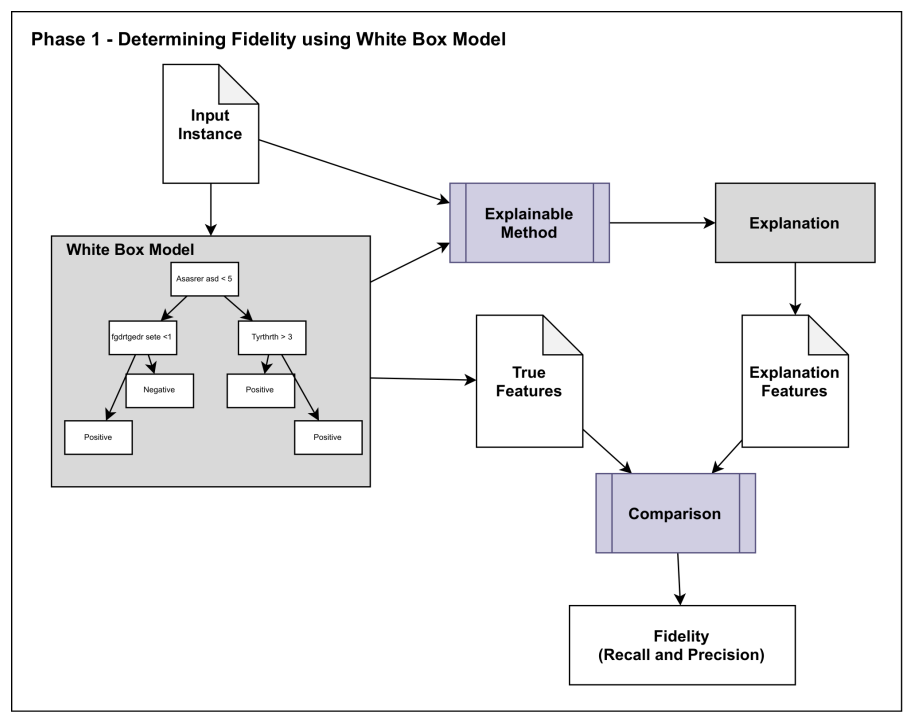}
    }
    \caption{A summary of Phase 1 of the evaluation approach. For a single input instance, an explanation is created from a white box model, and an explanation is created by an explainable method. The features that impacted the prediction are extracted from the white box model (true features) and the features implied to be most important are extracted from the explanation (explanation features). The two sets of features are then compared to determine fidelity.}
    \label{fig:phase1}
\end{figure}

As such, the evaluation method in this phase is as follows (see figure~\ref{fig:phase1}):
\begin{enumerate}
    \item Generate a white box predictive model and determine the appropriate number of features needed to calculate recall $\mathcal{R}$ ($n$).
    \item Extract the set of $True~Features$ for input instance $x$.
    \item Generate $k$ explanations for input instance $x$, and compute the average feature weight for each feature. This step is taken to mitigate any instability present within the explanation. We choose $k=10$ in our work.
    \item Based on the absolute value of the weights for each feature, extract the top $n$ features from the explanation ($Explanation~Features$) and use equation~\ref{eq:recall} to calculate recall.
    \item Based on the absolute value of the weights for each feature, extract \textit{Explanation~Features} in the top quartile of the feature weight distribution from the explanation and use equation~\ref{eq:precision} to calculate precision.
\end{enumerate}

\subsection{Phase 2: White Box as Proxy for Black Box}\label{sec:phase-2}

The aim of the second phase is to develop a method to test the fidelity of explanations for a black box, using a white box model as a substitute to determine the appropriateness of the method and parameters used in the method. Specifically, the approach we proposed in a previous work will be refined and extended to better evaluate the fidelity of black box models~\citep{Velmurugan2021}. This approach is adapted from an ablation-based method used to test the internal fidelity of tabular and text data, which tests the correctness of the explanation.

The general approach of this method is to identify the features deemed to be important by the explanation, and perturb or remove these features from the original input. A prediction is then generated from this new input, and the original and new prediction are compared~\citep{Du2019,Fong2017,Nguyen2018,Samek2017}. When evaluating predictive models trained on image or text data, the relevant features can simply removed from the original input or blurred, and the degree of change in the prediction probability for the originally predicted class is used as an indicator of the correctness of the explanation~\citep{Du2019,Fong2017,Nguyen2018,Samek2017}. While this approach of ``removal'' of features is relatively simple in image or text data, this approach will not hold for tabular data, were "gaps" in the input are automatically imputed by the predictive model, or are otherwise treated as some improbable value, such as infinity. Thus, in a previous work~\citep{Velmurugan2021}, rather than removing features, we attempted to alter them through perturbation to reflect a value outside of the range of values considered to be relevant by the explanation.

However, there are a number of parameters that must first be determined in order to perturb the input. As these parameters cannot be extracted from a black box model, given its opaque nature, white box models can be used to identify the most appropriate values -- or at least provide a heuristic that can be applied when evaluating black box models. First is the identification of the specific feature to be perturbed. Given that a typical feature attribution explanation does not specifically highlight a set of features as the most relevant, only providing the importance of all features, this relevant set must be extracted from the explanation. The feature importance weight provided by the explanation can be used for this purpose. Features with a weight higher than a specific point in the feature weight distribution can be considered to be the ``most relevant'' features, though, this threshold between ``most relevant'' and less relevant must first be identified.

\begin{figure}[h!]
    \resizebox{\columnwidth}{!} {
    \includegraphics[scale = 0.5]{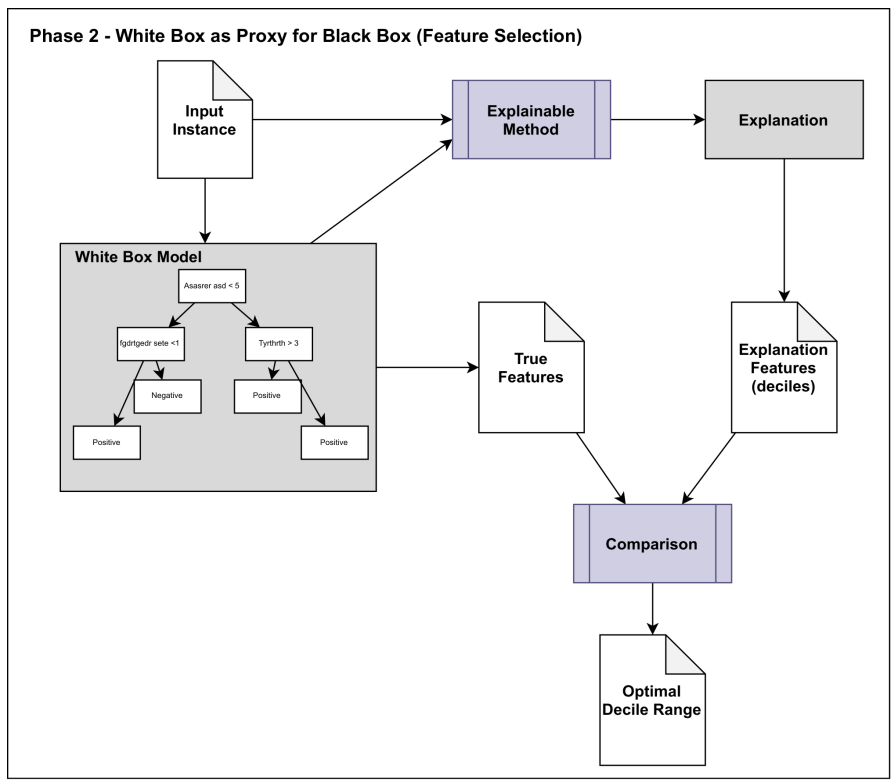}
    }
    \caption{A summary of the approach taken to identify features for perturbation in Phase 2. Once the true features from a white box model are identified, and the features in the explanation are sorted into deciles based on the distribution of feature weights provided by the explanation. Every possible range of deciles from only the first to the first to the ninth are compared against the true features to identify the most optimal range of deciles, which maximises both feature recall and feature precision. }
    \label{fig:phase2a}
\end{figure}

One way to determine this threshold is through division of the feature weight distribution, much as with Phase 1. However, rather than quartiles, the feature weight distribution is divided into deciles in Phase 2, to take a more fine--grained approach to determining the most appropriate and relevant set of features to use. The simplest way to identify the most appropriate number of deciles to form the "most relevant" set of features. Once again, precision and recall will be calculated for each possibility, starting from only the first decile (features with the largest feature weights) to using features from the first to the ninth deciles. As the set of explanation features used to calculate both recall and precision are identical here, the F1 score can be calculated, and used to determine the most optimal decile range (figure~\ref{fig:phase2a}).

A second parameter to consider is the feature value range considered relevant by the explanation. While this range is provided as part of the explanation by some explainable methods, such as LIME, other explainable methods, including SHAP, provide only a feature weight for each feature. In these cases, the appropriate range will need to be from the explanation based only on feature weights. An appropriate way to find such a range would be to identify the range of feature values that provide a similar feature weight to the input under explanation, and, as such, what constitutes a "similar feature weight" must also be defined. Given that feature weights are often at different scales between different explainable methods, or even between different datasets when applying the same explainable method (for example, in regression problems where the prediction targets are at different scales), the most expedient way to determine similarity may be to identify feature weights that are no less or no more than a certain percentage of the feature weight for the instance being explained. For example, feature weights that are no more than 10\% smaller or 10\% larger than the feature weight for the instance being explained, in essence creating a sliding scale bin centred on the initial feature weight.

This then raises the question of what percentage of the feature weight provides the most appropriate bin size. Once again, comparing the results of the search to the white box model can provide a heuristic that can be applied to a black box model. Once the feature weight bin has been chosen, all of the feature values that generate a feature weight that fall into that bin can be identified, and the minimum and maximum of those feature values is considered to be the threshold provided by the explanation. Once the explanation-derived thresholds have been extracted, the minimum and maximum values of that range can be compared to the minimum and maximum threshold values for that feature in the white box mode. In the case of a decision tree, the maximum and minimum possible feature values for that feature to follow the decision path can be compared to the explanation-derived threshold. The optimal bin size would be the percentage of the initial feature weight that minimises the distance between the explanation-derived thresholds and the white box model's thresholds across all features. 

\begin{figure}[h!]
    \resizebox{\columnwidth}{!} {
    \includegraphics[scale = 0.5]{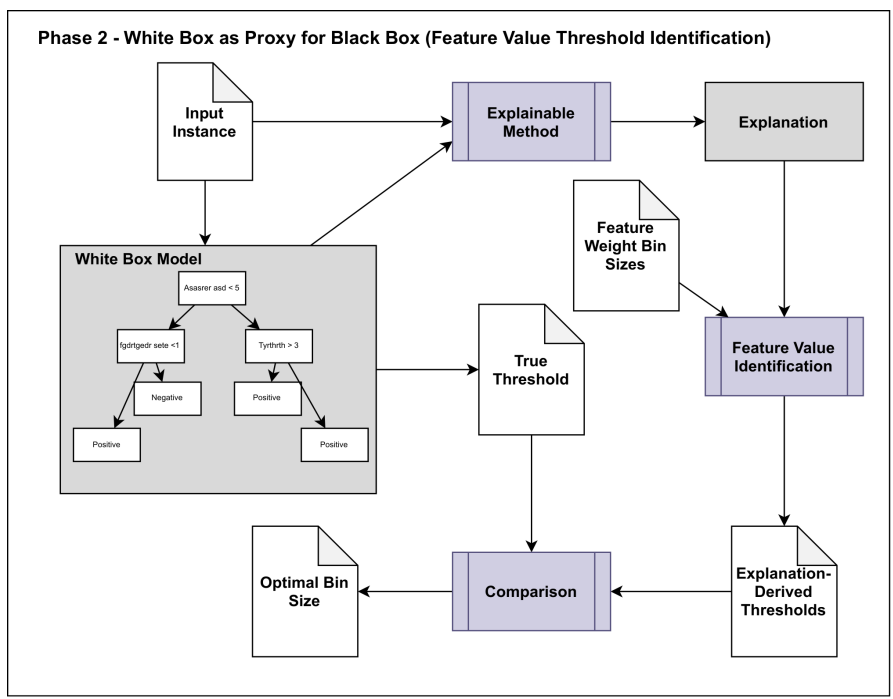}
    }
    \caption{A summary of the feature value threshold identification approach in Phase 2. Once the true thresholds for each feature are derived from the white box model, similar thresholds are also extracted from the explanation. These thresholds are based on the range of feature values that produce feature weights falling within a bin of feature weights similar to the feature weight for the instance under explanation. The true and explanation-derived thresholds are then compared to find the optimal bin size, which would minimise the distance between thresholds from the two sources.}
    \label{fig:phase2b}
\end{figure}

In this phase, some heuristic or general value will be extracted for these parameters, to be used as part of the fidelity evaluation method for black box models. The evaluation method (described in section~\ref{sec:phase-3}) can then be applied to explanations of the white box model, and compared against the Phase 1 results to determine the validity of the method.
\subsection{Phase 3: Determining Fidelity using Black Box Model}\label{sec:phase-3}

The identified heuristics for parameters discovered in phase 2 will be applied during this phase to evaluate explanations for a black box model.As noted, the general approach for the fidelity evaluation method proposed in this work mirrors ablation-based methods used for text and image data, wherein the data is altered through ``removal'' of relevant features and the changes in prediction probability for the original and altered inputs is compared~\citep{Du2019,Fong2017,Nguyen2018,Samek2017}. Two major changes are made to adapt this method for tabular data. 

Firstly, as noted in section~\ref{sec:phase-2}, inputs must be altered through perturbation rather than ablation. In text data, the ablation approach is simply applied by removing features (typically words, phrases or sentences) identified as important by the explanation. When perturbing the data, however, the nature of the change becomes less absolute. Moreover, there are two possible methods of perturbations. Given the relevant range of feature values, the feature value of the input can be altered to something inside this range, or to a value outside of this range. The former method would test the relevance of the range, where relatively small changes to the model output would indicate that the feature value range is truly relevant and a larger change would indicate that irrelevant values are included in the range. We term this \textit{explanation-supporting fidelity}. 

The latter method of perturbing outside of the range can be used to the test the firmness of the boundaries indicated by the range. That, is, do feature values outside of the relevant range produce a vastly different result to the original? Some explainable methods, such as Anchor, limit the ranges provided by the explanation in order to maximise the precision of the explanation, based on a threshold of correctness within that range~\citep{Riberio2018}. Thus, values outside of the range provided by the explanation may produce similar results, but not often enough to fall within the minimum required threshold. In order to test this firmness of the boundary, we can perturb the value of a feature to that just outside of the boundary indicated by the explanation, and, if the altered input produces a significantly different output than the original, this indicates that the boundary is firm. We term this \textit{explanation-contrary fidelity}.

Secondly, the ablation approach applied in literature was typically applied to classification problems. As such, the method determining fidelity was to calculate the differences in prediction probability for the originally predicted class for the two inputs. We maintain this method for classification problems, but compare changes in prediction in regression problems.

\begin{figure}[h!]
    \resizebox{\columnwidth}{!} {
    \includegraphics[scale = 0.5]{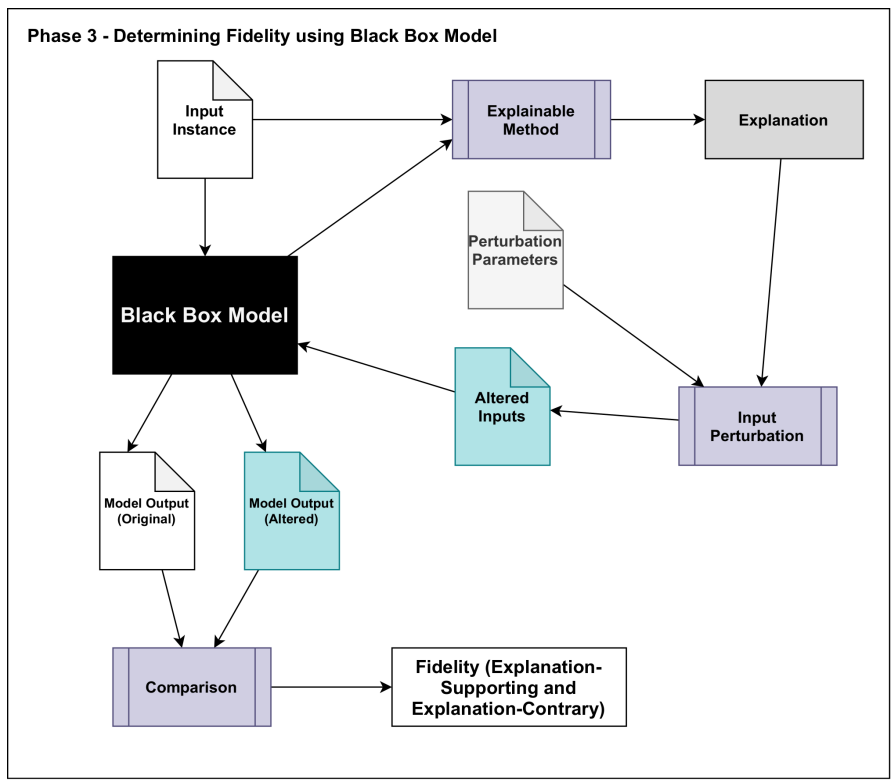}
    }
    \caption{A summary of Phase 3 of the evaluation approach. Using a black box model, a prediction is generated for each instance, as well as an explanation from a post-hoc explanation method. Perturbations of the original input are created based on the explanation (using the perturbation parameters from the previous phase) and model outputs are generated for these perturbed inputs. The model outputs for the original and perturbed instances are compared to determine fidelity. }
    \label{fig:phase3}
\end{figure}

As such, the method of evaluation is as follows (see figure~\ref{fig:phase3}):
\begin{enumerate}
    \item Record the output of the predictive model ($Y(x)$) for the inital input $x$.
    \item Generate $k$ explanations for the input $x$, and compute the average feature weight for each feature. This step is taken to mitigate any instability present within the explanation.
    \item Based on the absolute value of the weights for each feature, identify all features that fall into the relevant decile range (this range is determined from the results of Phase 2).
    \item Extract the relevant feature value range for all features identified in the previous step. For methods that provide only feature weights as explanation, we use the optimised bin size determined during Phase 2.
    \item For all features identified in step 3, replace the value of that feature with a randomly generated value (from within the range to evaluate explanation-supporting fidelity and just outside of the range for explanation-contrary fidelity), creating the altered input $x'$.
	\item Record the output of the predictive model ($Y(x')$) for the altered input $x'$.
\end{enumerate}

This method was repeated ten times, and the MAPE of the differences between $Y(x)$ and $Y(x')$ was used to determine fidelity. To calculate explanation-contrary fidelity ($\mathcal{C}$), the MAPE was applied as:
\begin{equation}\label{eq:contrary-fidelity}
\mathcal{C} = \frac{\sum^{|X'|}_1\frac{|Y(x)-Y(x')|}{Y(x)}}{|X'|}
\end{equation}
where: \vspace*{-.5\baselineskip}
\begin{itemize}	
\item $x$ = original input for the process instance 
\item $X'$ = Set of perturbations for $x$ and $x'\in X'$
\item $Y(x)$ = Model output given input $x$
\item $Y(x')$ = Model output given input $x'$
\end{itemize}

\leavevmode\newline
 
For explanation-supporting fidelity ($\mathcal{S}$), this was applied as:
\begin{equation}\label{eq:supporting-fidelity}
\mathcal{S} = 1-\frac{\sum^{|X'|}_1\frac{|Y(x)-Y(x')|}{Y(x)}}{|X'|}
\end{equation}
where: \vspace*{-.5\baselineskip}
\begin{itemize}	
\item $x$ = original input for the process instance 
\item $X'$ = Set of perturbations for $x$ and $x'\in X'$
\item $Y(x)$ = Model output given input $x$
\item $Y(x')$ = Model output given input $x'$
\end{itemize}

Both metrics are applied at the instance-level, but can be averaged out over the dataset.
\section{Design of Experiments}
\label{sec:setup}
\subsection{Datasets}
Six open-source datasets were sourced from the UCI Machine Learning repository to conduct the evaluation described above (see table~\ref{tab:data-characteristics} for a summary of these datasets). All of the chosen datasets are relatively well-known and simple. Three of the datasets used were classification problems: a dataset to diagnose the malignance of breast cancer tumours\footnote{https://archive.ics.uci.edu/ml/datasets/Breast+Cancer+Wisconsin+(Diagnostic)}; a dataset to diagnose diabetes\footnote{https://archive.ics.uci.edu/ml/datasets/pima+indians+diabetes}; and the adult income prediction dataset\footnote{https://archive.ics.uci.edu/ml/datasets/Adult}. All these of these datasets were balanced through downsampling to ensure class parity. The remaining three are well-known regression prediction problems: house price prediction\footnote{https://archive.ics.uci.edu/ml/machine-learning-databases/housing/}; bike rental prediction\footnote{https://archive.ics.uci.edu/ml/datasets/Bike+Sharing+Dataset}; and student result prediction\footnote{https://archive.ics.uci.edu/ml/datasets/Student+Performance}. All but the Adult Income and Student Results datasets only had numerical features and were used as-is. As the other two also had categorical features, these features were one-hot encoded and the numerical features were used as-is. Once processed as necessary, the datasets were split into training and testing datasets and used to train the predictive models. F1-scores were used to measure the accuracy of the classification models, and the Mean Absolute Error (MAE) and Mean Absolute Percentage Error (MAPE) were used to measure the accuracy of the regression models.

\begin{table}[h]
\centering
\caption{Characteristics of datasets used}
\label{tab:data-characteristics}
\begin{tabular}{|l|c|r|r|}
\hline
\multicolumn{1}{|c|}{\textbf{Dataset}} &
  \textbf{Type} &
  \textbf{\# Features} &
  \textbf{\begin{tabular}[c]{@{}c@{}}\# Training\\ Instances\end{tabular}} \\ \hline
Diabetes             & Classification & 8   & 375   \\ \hline
Breast Cancer        & Classification & 30  & 296   \\ \hline
Adult Income         & Classification & 104 & 10977 \\ \hline
Boston House Pricing & Regression     & 13  & 354   \\ \hline
Bike Sharing         & Regression     & 12  & 12165 \\ \hline
Student Results      & Regression     & 58  & 454   \\ \hline
\end{tabular}
\end{table}

\subsection{Predictive Models}
Two types of predictive models were used to conduct the evaluations. As the first two phases of the evaluation require a fully-transparent white box model, a simple decision tree model was created for each dataset. For the black-box model evaluation in Phase 3, XGBoost models were created for each dataset. XGBoost models are ensembles of decision tree models, each building on the previous decision tree to improve its decision-making. As such, both predictive models used are decision-tree-based and belong to the same class of predictive models. A description of the trained models is provided in tables~\ref{tab:decision-tree-characteristics} and~\ref{tab:xgb-characteristics}.

\begin{table}[h!]
\centering
\caption{Characteristics of trained decision tree models.}
\label{tab:decision-tree-characteristics}
\begin{tabular}{|c|r|r|r|r|r|r|}
\hline
\textbf{Dataset} &
  \multicolumn{1}{c|}{\textbf{Accuracy (F1)}} &
  \multicolumn{1}{c|}{\textbf{MAE}} &
  \multicolumn{1}{c|}{\textbf{MAPE}} &
  \multicolumn{1}{c|}{\textbf{\begin{tabular}[c]{@{}c@{}}Max \\ Depth\end{tabular}}} &
  \multicolumn{1}{c|}{\textbf{\begin{tabular}[c]{@{}c@{}}Split \\ Nodes\end{tabular}}} &
  \multicolumn{1}{c|}{\textbf{\begin{tabular}[c]{@{}c@{}}Leaf \\ Nodes\end{tabular}}} \\ \hline
Diabetes                                                       & 0.6901 & -       & -      & 5  & 8    & 9    \\ \hline
Breast Cancer                                                  & 0.8833 & -       & -      & 5  & 6    & 7    \\ \hline
Adult Income                                                   & 0.8225 & -       & -      & 17 & 54   & 55   \\ \hline
\begin{tabular}[c]{@{}c@{}}Boston House\\ Pricing\end{tabular} & -      & 2.9690  & 0.1419 & 19 & 343  & 344  \\ \hline
Bike Sharing                                                   & -      & 33.0928 & 0.3429 & 19 & 24   & 25   \\ \hline
Student Results                                                & -      & 0.8917  & 0.0822 & 28 & 3515 & 3516 \\ \hline
\end{tabular}
\end{table}

\begin{table}[h]
\centering
\caption{Characteristics of trained XGBoost models.}
\label{tab:xgb-characteristics}
\begin{tabular}{|c|r|r|r|r|r|r|}
\hline
\textbf{Dataset} &
  \multicolumn{1}{c|}{\textbf{Accuracy (F1)}} &
  \multicolumn{1}{c|}{\textbf{MAE}} &
  \multicolumn{1}{c|}{\textbf{MAPE}} &
  \multicolumn{1}{c|}{\textbf{\begin{tabular}[c]{@{}c@{}}Num\\ Trees\end{tabular}}} &
  \multicolumn{1}{c|}{\textbf{\begin{tabular}[c]{@{}c@{}}Split \\ Nodes\end{tabular}}} &
  \multicolumn{1}{c|}{\textbf{\begin{tabular}[c]{@{}c@{}}Leaf \\ Nodes\end{tabular}}} \\ \hline
Diabetes                                                       & 0.7215 & -       & -      & 100 & 4767  & 4867  \\ \hline
Breast Cancer                                                  & 0.9672 & -       & -      & 100 & 1178  & 1278  \\ \hline
Adult Income                                                   & 0.8404 & -       & -      & 100 & 100   & 200   \\ \hline
\begin{tabular}[c]{@{}c@{}}Boston House\\ Pricing\end{tabular} & -      & 3.0929  & 0.1541 & 100 & 100   & 200   \\ \hline
Bike Sharing                                                   & -      & 37.2251 & 0.9636 & 100 & 100   & 200   \\ \hline
Student Results                                                & -      & 0.9110  & 0.0774 & 100 & 10540 & 10640 \\ \hline
\end{tabular}
\end{table}

\subsection{Explainable Methods}
LIME and SHAP, two popular feature attribution methods are evaluated in this work. Both are typical local feature attribution methods that, for a single instance, provide an explanation by highlighting how important each feature in the input was to the final prediction (as explained in section~\ref{sec:background}). As such, given their popularity and representativeness of this class of explaninable methods, they were chosen for evaluation. SHAP includes a number of explainable mechanisms that are optimised for different classes of predictive models, as well as a fully model-agnostic variant. In this work, as only tree-based models are used, TreeSHAP is used for evaluation. 

As part of the three-phase approach outlined in section~\ref{sec:approach}, explanations were first generated from both explainable methods for the decision tree model, and these explanations were compared directly against the workings of the decision tree. In the second phase, the optimal number of deciles to draw features from for perturbation were searched for for both LIME and SHAP. The appropriate bin scale to choose similar feature weights and discover the associated feature values was also required for SHAP. Specifically, the percentage of the initial SHAP value required to identify relevant feature values was searched for during Phase 2. Finally, once the parameters were finalised, the both explainable methods were evaluated when explaining the black box XGBoost models.

All of the code, datasets used for evaluation and results of evaluation are available at: \url{https://git.io/JZdVR}.
\section{Results and Analysis}~\label{sec:results}
\subsection{Results and Observations}\label{observations}
\subsubsection{Phase 1: Determining Fidelity using White Box Model}
Results from Phase 1 of the evaluation development approach, where the explanations are evaluated directly against the white box model, show that LIME and SHAP are generally comparable in fidelity (table~\ref{tab:phase-1-result}). Though LIME generally has higher recall on dataset with a smaller amount of features (the Diabetes, Breast Cancer and Boston House Pricing datasets), the differences between the two datasets in recall is generally small. The two exceptions to this are the Adult Income and Student Results datasets, where LIME's recall is significantly lower. It's also noteworthy that these two datasets have the greatest amount of features out of the two (Adult Income has 104 features and Student Results has 58 features). Interestingly, this does not hold true for Precision. Both LIME and SHAP are generally comparable in this regard, and the number of features does not appear to affect LIME's results.

\begin{table}[h!]
\centering
\caption{Fidelity Results from Phase 1}
\label{tab:phase-1-result}
\begin{tabular}{|c|c|r|r|r|r|}
\hline
\multicolumn{2}{|c|}{\multirow{2}{*}{\textbf{Dataset}}}    & \multicolumn{2}{c|}{\textbf{Recall}}                  & \multicolumn{2}{c|}{\textbf{Precision}}               \\ \cline{3-6} 
\multicolumn{2}{|c|}{}                                     & \multicolumn{1}{c|}{LIME} & \multicolumn{1}{c|}{SHAP} & \multicolumn{1}{c|}{LIME} & \multicolumn{1}{c|}{SHAP} \\ \hline
\multirow{3}{*}{\textbf{Classification}} & Diabetes        & \textbf{0.9982}           & 0.9396                    & 0.8937                    & \textbf{0.9722}           \\ \cline{2-6} 
                                         & Breast cancer   & \textbf{0.8545}           & 0.8517                    & \textbf{1.000}            & \textbf{1.000}            \\ \cline{2-6} 
                                         & Adult Income    & 0.7563                    & \textbf{0.9917}           & \textbf{0.9584}           & 0.8495                    \\ \hline
\multirow{3}{*}{\textbf{Regression}} & \begin{tabular}[c]{@{}c@{}}Boston House Pricing\end{tabular} & \textbf{0.8847} & 0.8821 & 0.8169 & \textbf{0.9320} \\ \cline{2-6} 
                                         & Bike Sharing    & 0.8031                    & \textbf{0.8216}           & \textbf{1.000}            & 0.9991                    \\ \cline{2-6} 
                                         & Student Results & 0.8022                    & \textbf{0.9904}           & 0.9402                    & \textbf{0.9949}           \\ \hline
\end{tabular}
\end{table}

The trade off between recall and precision is apparent in these results. For four out of the six datasets, that the explanation method that had the higher recall was not the explanation method that had produced the more precise explanations. The exceptions to this are the Breast Cancer and Student Results datasets.
\subsubsection{Phase 2: White Box as Proxy for Black Box}
Three parameters were being trialled during Phase 2: (i) the number of deciles holding the most relevant features in LIME explanations (table~\ref{tab:lime-param}) and (ii) the number of deciles holding the most relevant features in SHAP explanations and (iii) the optimised bin size for SHAP values to allow us to identify the relevant feature value range (table~\ref{tab:shap-param}).

\begin{table}[h!]
\centering
\caption{Optimal fidelity parameters for SHAP}
\label{tab:shap-param}
\begin{tabular}{|c|c|c|c|c|c|c|}
\hline
\multicolumn{2}{|c|}{\textbf{Dataset}} &
  \textbf{\begin{tabular}[c]{@{}c@{}}Optimal \\ Decile\end{tabular}} &
  \textbf{Recall} &
  \textbf{Precision} &
  \textbf{F1} &
  \textbf{\begin{tabular}[c]{@{}c@{}}Optimal\\ Bin Size\end{tabular}} \\ \hline
\multirow{3}{*}{\textbf{Classification}} & Breast Cancer   & 8 & 0.7034 & 0.6064 & 0.7465 & 1.0 \\ \cline{2-7} 
                                & Diabetes        & 8 & 0.8436 & 0.7955 & 0.8189 & 0.1 \\ \cline{2-7} 
                                & Adult Income    & 9 & 0.7349 & 0.8180 & 0.7742 & 0.5 \\ \hline
\multirow{3}{*}{\textbf{Regression}} &
  \begin{tabular}[c]{@{}c@{}}Boston House \\ Prices\end{tabular} &
  9 &
  0.7239 &
  0.7291 &
  0.7265 &
  0.1 \\ \cline{2-7} 
                                & Bike Sharing    & 9 & 0.5932 & 0.9078 & 0.7175 & 0.1 \\ \cline{2-7} 
                                & Student Results & 9 & 0.7380 & 0.9808 & 0.8423 & 0.7 \\ \hline
\end{tabular}
\end{table}

\begin{table}[h!]
\centering
\caption{Optimal fidelity parameters for LIME}
\label{tab:lime-param}
\begin{tabular}{|c|c|c|c|c|c|}
\hline
\multicolumn{2}{|c|}{\textbf{Dataset}} &
  \textbf{\begin{tabular}[c]{@{}c@{}}Optimal \\ Decile\end{tabular}} &
  \textbf{Recall} &
  \textbf{Precision} &
  \textbf{F1} \\ \hline
\multirow{3}{*}{\textbf{Classification}} &
  Breast Cancer &
  7 &
  0.5932 &
  0.6949 &
  0.6383 \\ \cline{2-6} 
 & Diabetes        & 7 & 0.7240 & 0.7515 & 0.7375 \\ \cline{2-6} 
 & Adult Income    & 8 & 0.5140 & 0.7155 & 0.5982 \\ \hline
\multirow{3}{*}{\textbf{Regression}} &
  \begin{tabular}[c]{@{}c@{}}Boston House \\ Prices\end{tabular} &
  9 &
  0.6721 &
  0.6449 &
  0.6583 \\ \cline{2-6} 
 & Bike Sharing    & 9 & 0.4648 & 0.8137 & 0.5916 \\ \cline{2-6} 
 & Student Results & 9 & 0.6350 & 0.8362 & 0.7317 \\ \hline
\end{tabular}
\end{table}

There is a marked difference in optimal parameters for classification models and regression models. For both LIME and SHAP. Classification models generally require fewer deciles of features than regression models, and LIME sorts the relevent features into fewer deciles than SHAP. Similarly, the optimal bin size varies greatly when explaining classification models, but has a smaller range when explaining regression models. All but the last decile of features produces the highest F1 score when explaining regression models, while 8 deciles of features are generally the most appropriate for SHAP and 7 deciles of features are generally the most appropriate for LIME when explaining classification models. The only exception to this is in the Adult Income dataset, where 9 deciles of features is the most appropriate for SHAP and 8 deciles of features is the most appropriate for LIME.

Unlike in Phase 1, there is a clear difference between the F1 scores of LIME and SHAP when considering the optimal deciles of explanations. For all models explained, both classification and regression, SHAP has the highest F1 scores. However, it is also clear that the increase in F1 scores comes at the cost of precision, as precision never rises beyond 0.84 when using LIME, and only rises past 0.9 when using one of two datasets with SHAP (though it also does not fall below 0.6 at any time, so precision remains relatively high). This is in stark contract to Phase 1, where precision only fell below 0.85 for two combinations of dataset and explainable method. 

Optimal parameters to be used with the method described in section~\ref{sec:phase-2} were identified using the optimal values of quartiles and bin scales for each dataset in tables~\ref{tab:shap-param} and~\ref{tab:lime-param}. Given the clear distinction between optimal values for classification and regression models, separate values are required for each. As such, for classification problems, the mean of the optimal bin sizes for each type of prediction problem is used. That is, 50\% of the original SHAP value is used to create the SHAP value bins to identify feature values that produce SHAP values that fall into that bin. For regression, 30\% of the original SHAP value is used. Features in the top nine deciles are used as features to be perturbed for regression problems for both LIME and SHAP, whereas, for classification problems, 7 deciles are used for LIME and 8 are used for SHAP.

Using these parameters, explanation-supporting and explanation-contrary fidelity are both tested for the white box decision tree models (table~\ref{tab:phase-2-result}). Once again, barring a few notable exceptions, LIME and SHAP have similar or identical fidelity. Explanation-supporting fidelity, in particular, is quite similar or identical, and always high. Explanation-contrary fidelity, by contrast, is almost always low. Moreover, LIME has poorer explanation-contrary fidelity than SHAP for 
the Breast Cancer and Adult Income datasets. In contrast to the classification models, the explanation-supporting fidelity for regression datasets is uniformly perfect (1 for all datasets) and explanation-contrary fidelity is uniformly poor (0 for all datasets), for both LIME and SHAP.

\begin{table}[h!]
\centering
\caption{Fidelity Results from Phase 2}
\label{tab:phase-2-result}
\begin{tabular}{|c|c|r|r|r|r|}
\hline
\multicolumn{2}{|c|}{\multirow{2}{*}{\textbf{Dataset}}} & \multicolumn{2}{c|}{\textbf{\begin{tabular}[c]{@{}c@{}}Explanation\\ Supporting\end{tabular}}} &
  \multicolumn{2}{c|}{\textbf{\begin{tabular}[c]{@{}c@{}}Explanation\\ Contrary\end{tabular}}} \\ \cline{3-6} 
\multicolumn{2}{|c|}{} & \multicolumn{1}{c|}{LIME} & \multicolumn{1}{c|}{SHAP} & \multicolumn{1}{c|}{LIME} & \multicolumn{1}{c|}{SHAP} \\ \hline
\multirow{3}{*}{\textbf{Classification}} & Diabetes & 0.9306 & \textbf{0.9824} & \textbf{0.4648} & 0.4480 \\ \cline{2-6} 
 & Breast cancer & 0.9289 & \textbf{1.000} & 0.3780 & \textbf{0.6701} \\ \cline{2-6} 
 & Adult Income & \textbf{0.9036} & 0.8032 & 0.1592 & \textbf{0.4103} \\ \hline
\multirow{3}{*}{\textbf{Regression}} & Boston House Pricing & \textbf{1.000} & \textbf{1.000} & \textbf{0.000} & \textbf{0.000} \\ \cline{2-6} 
 & Bike Sharing & \textbf{1.000} & \textbf{1.000} & \textbf{0.000} & \textbf{0.000} \\ \cline{2-6} 
 & Student Results & \multicolumn{1}{r|}{\textbf{1.000}} & \multicolumn{1}{r|}{\textbf{1.000}} & \multicolumn{1}{r|}{\textbf{0.000}} & \multicolumn{1}{r|}{\textbf{0.000}} \\ \hline
\end{tabular}
\end{table}

\subsubsection{Phase 3: Determining Fidelity using Black Box Model}

An interesting pattern emerges in the Phase 3 results (table~\ref{tab:phase-3-result}) for classification models. Once again, explanation-supporting fidelity is perfect and explanation-contrary fidelity is 0 for all regression models. However, the results for classification models are more varied. In particular, while SHAP explanations appear to have better explanation-supporting fidelity, LIME's explanations have significantly firmer boundaries (explanation-contrary fidelity), though explanation contrary fidelity is generally low. It is also worth noting that explanation-supporting fidelity is highly similar for the Diabetes and Breast Cancer datasets, but LIME's explanation-supporting fidelity significantly lower than that of SHAP for the Adult Income dataset.

\begin{table}[h]
\centering
\caption{Fidelity Results from Phase 3}
\label{tab:phase-3-result}
\begin{tabular}{|c|c|r|r|r|r|}
\hline
\multicolumn{2}{|c|}{\multirow{2}{*}{\textbf{Dataset}}} & \multicolumn{2}{c|}{\textbf{\begin{tabular}[c]{@{}c@{}}Explanation\\ Supporting\end{tabular}}} &
  \multicolumn{2}{c|}{\textbf{\begin{tabular}[c]{@{}c@{}}Explanation\\ Contrary\end{tabular}}} \\ \cline{3-6}
\multicolumn{2}{|c|}{} & \multicolumn{1}{c|}{LIME} & \multicolumn{1}{c|}{SHAP} & \multicolumn{1}{c|}{LIME} & \multicolumn{1}{c|}{SHAP} \\ \hline
\multirow{3}{*}{\textbf{Classification}} & Diabetes & 0.8105 & \textbf{0.8984} & \textbf{0.5023} & 0.1169 \\ \cline{2-6} 
 & Breast cancer & 0.9810 & \textbf{0.9968} & \textbf{0.1125} & 0.0076 \\ \cline{2-6} 
 & Adult Income & 0.6730 & \textbf{0.9816} & \textbf{0.3702} & 0.1176 \\ \hline
\multirow{3}{*}{\textbf{Regression}} & Boston House Pricing & \textbf{1.000} & \textbf{1.000} & \textbf{0.000} & \textbf{0.000} \\ \cline{2-6} 
 & Bike Sharing & \textbf{1.000} & \textbf{1.000} & \textbf{0.000} & \textbf{0.000} \\ \cline{2-6} 
 & Student Results & \textbf{1.000} & \textbf{1.000} & \textbf{0.000} & \textbf{0.000} \\ \hline
\end{tabular}
\end{table}

The results for the classification models are also generally poorer in this phase than in Phase 2, with some scattered exception. LIME's explanation-contrary fidelity is higher for the Diabetes and Adult Income datasets. LIME also has higher explanation-supporting fidelity for this latter dataset. Both LIME and SHAP also have higher explanation-supporting fidelity for the Breast Cancer dataset.
\subsection{Findings}\label{sec:analysis}

The most unexpected result of the three phases comes from Phases 2 and 3, where regression datasets had uniformly perfect explanation-supporting fidelity and explanation-contrary fidelity for both explanation methods, regardless of the predictive model used. This result is also perhaps the easiest to explain. Though technically a regression model, tree-based regressor models do not produce continuous outputs. Rather the models have a significant number of leaf nodes each of which produces a single output value. It is likely that the scope of the feature value perturbation made during Phases 2 and 3 were simply not enough to alter the final leaf node in the decision path of the models, and so produced the uniformly perfect and poor results for explanation supporting and explanation contrary respectively. Nevertheless, for regression problems using decision-tree based models, this high explanation-supporting fidelity suggests that the feature value distribution provided by the explanation is correct, though values outside of this feature range, but not distant, are still likely to provide identical results.

Another notable result is the performance of LIME in Phase 1. In Phase 1, LIME had higher recall for the Diabetes, Breast Cancer and House Prices datasets. Notably, these dataset also had significantly fewer features than the Adult Income and Student Results datasets (see table~\ref{tab:data-characteristics}). LIME is known to have poor stability~\citep{Visani2020}, which in turn appears to affect the fidelity of results here. This is apparent in the results, where LIME's recall for the Student Results dataset is significantly lower than SHAP's, and even lower sill for the Adult Income dataset. As the features in the data increase, LIME's recall of the most relevant features decreased. This is also apparent when determining explanation-contrary fidelity in Phase 2 and explanation-supporting fidelity in Phase 3 for the Adult Income dataset.

An exception to this rule appears to be the Bike Sharing dataset, where LIME had poorer recall than SHAP, though not significantly. Though this dataset has fewer features than both the Breast Cancer and House Pricing datasets, it also had the poorest model performance of the three regression models (table~\ref{tab:decision-tree-characteristics}). One past evaluation of LIME and SHAP found that explanation was linked to model performance~\citep{Yalcin2021}, and this is the likely explanation for this discrepancy. It is also worth noting that this dataset had the poorest SHAP recall out of all 6 datasets, though recall for both LIME and SHAP were not significantly different to that of the other datasets. While predictive model accuracy did seem to affect results, the impact appears not to have been significant. The Breast Cancer classifier and Student Results regressor were the most accurate in each category, but explanations for these datasets in have recall comparable to the other datasets. Precision, however, is both high and the more precise explainable method for both datasets is also the method that had higher recall.

\begin{figure}
    \centering
    \subfloat[Diabetes using Decision Tree]{
    \includegraphics[scale=0.4]{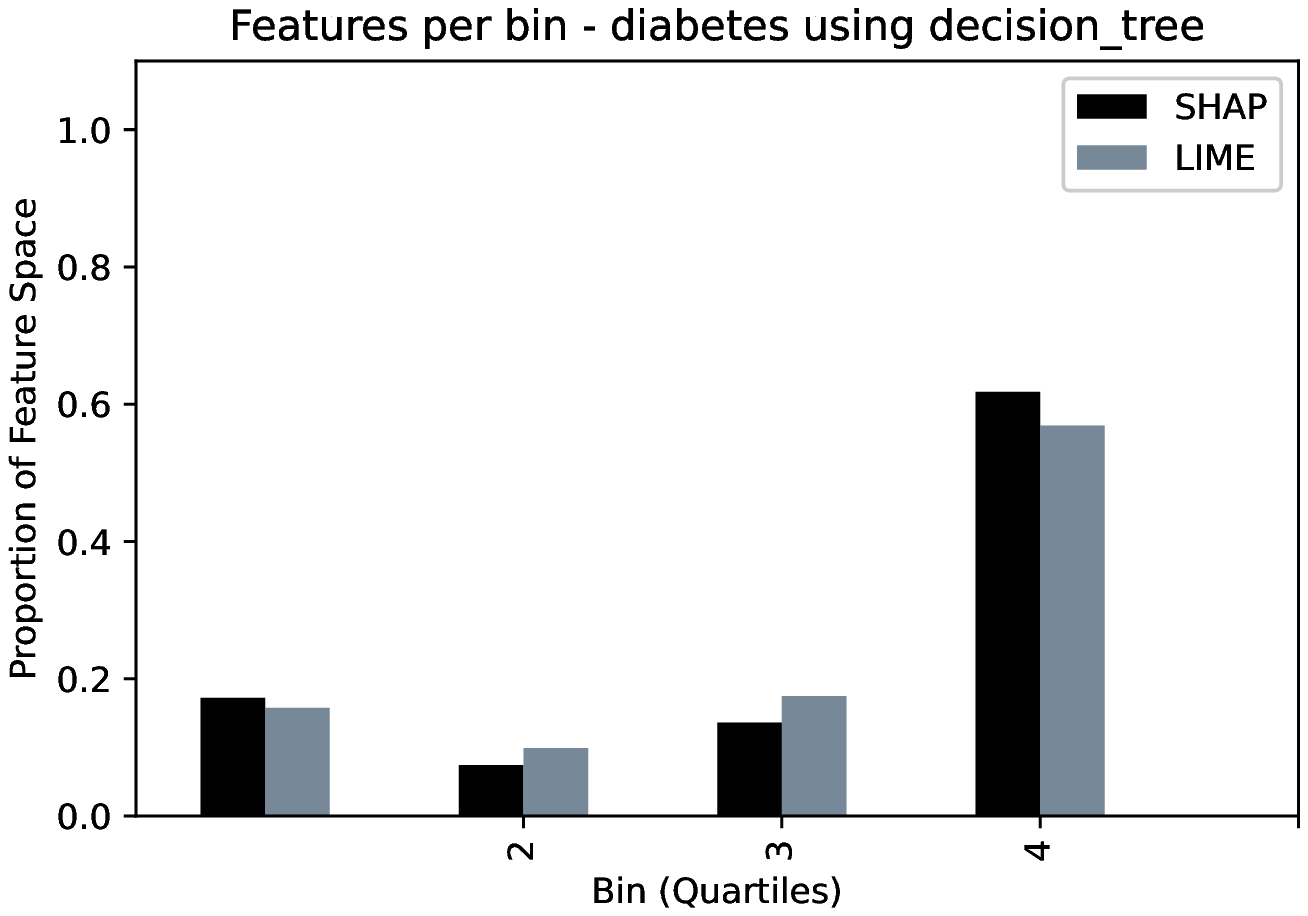}
    \includegraphics[scale=0.4]{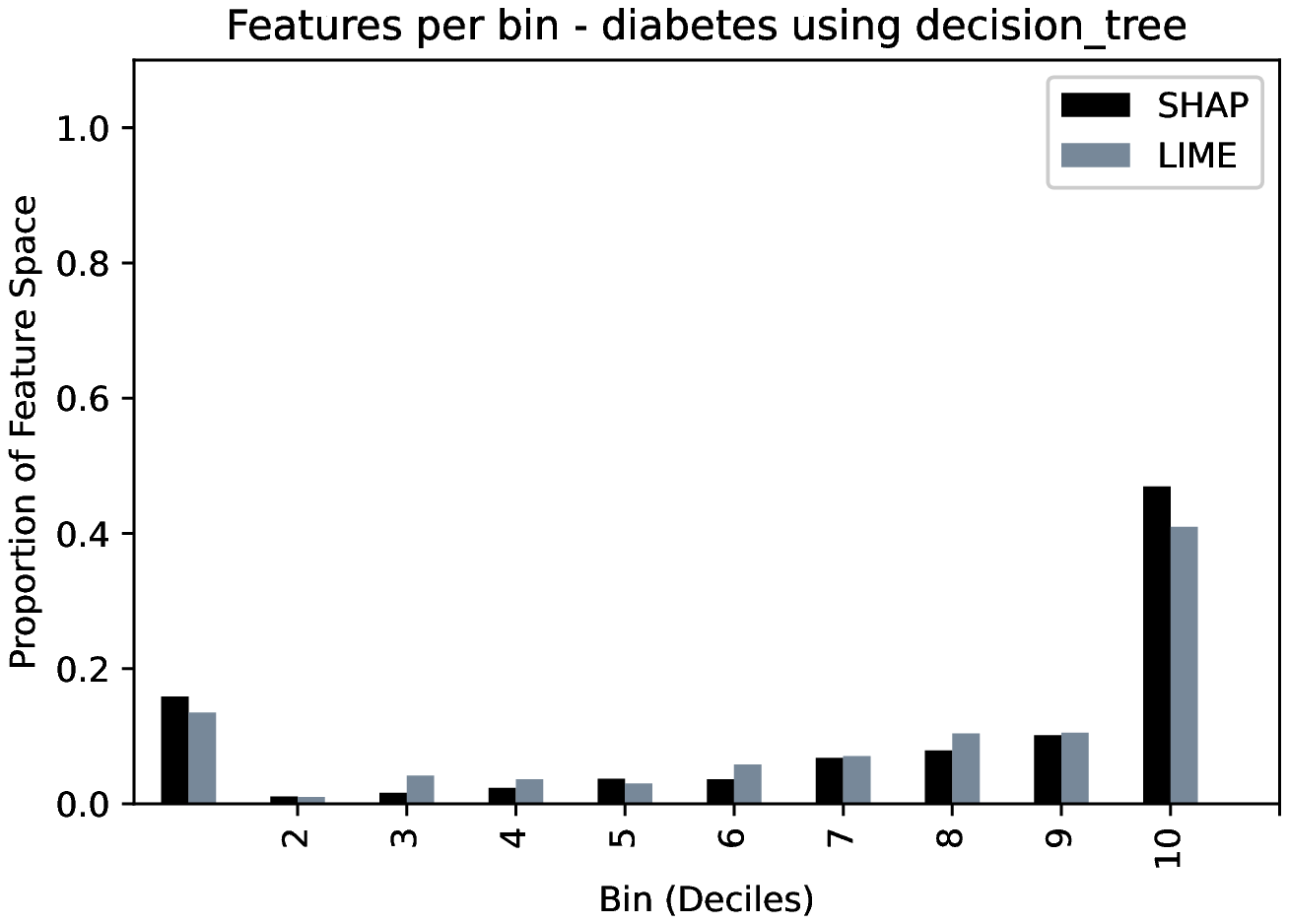}
    }\\
    \subfloat[Diabetes using XGBoost]{
    \includegraphics[scale=0.4]{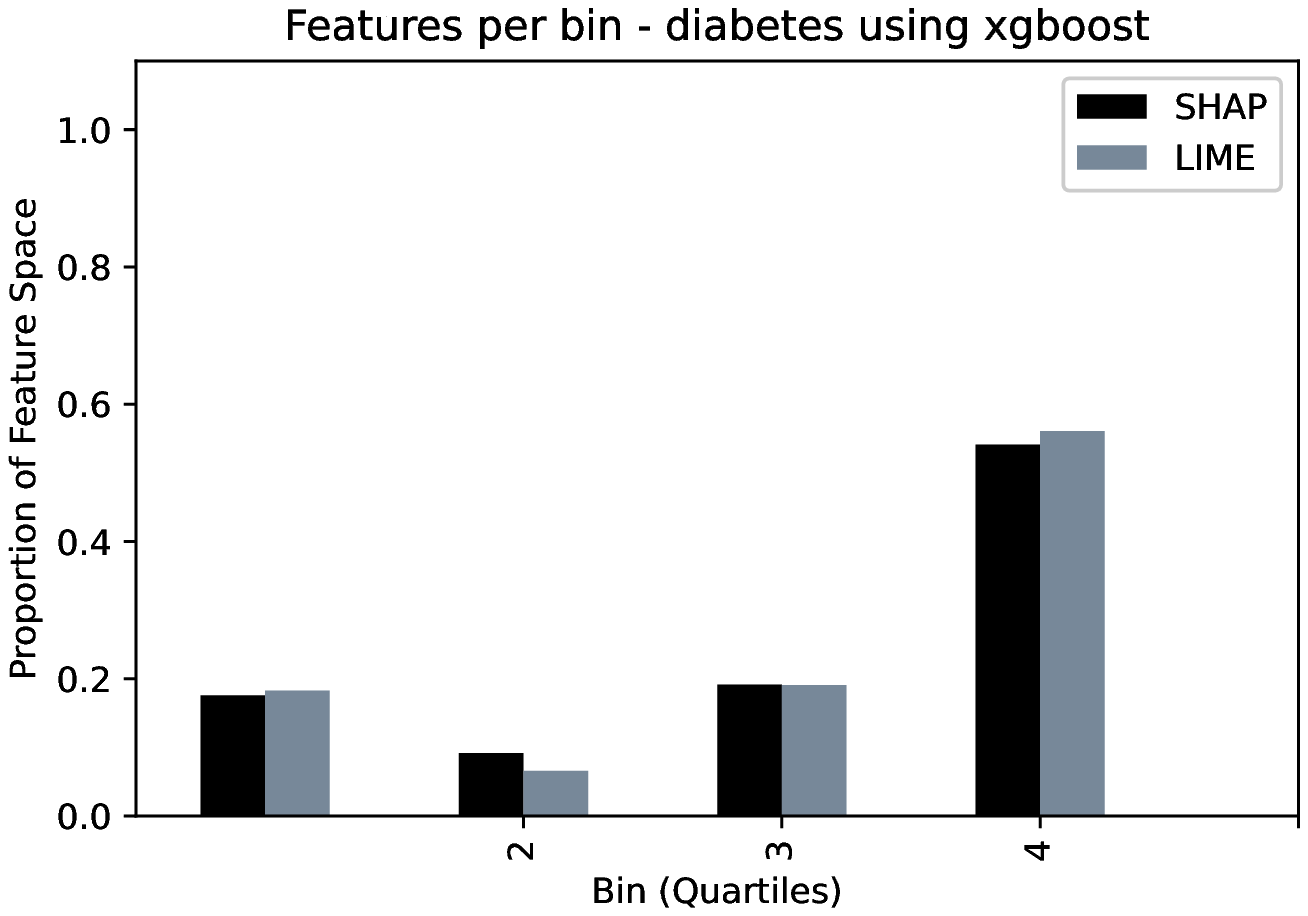}
    \includegraphics[scale=0.4]{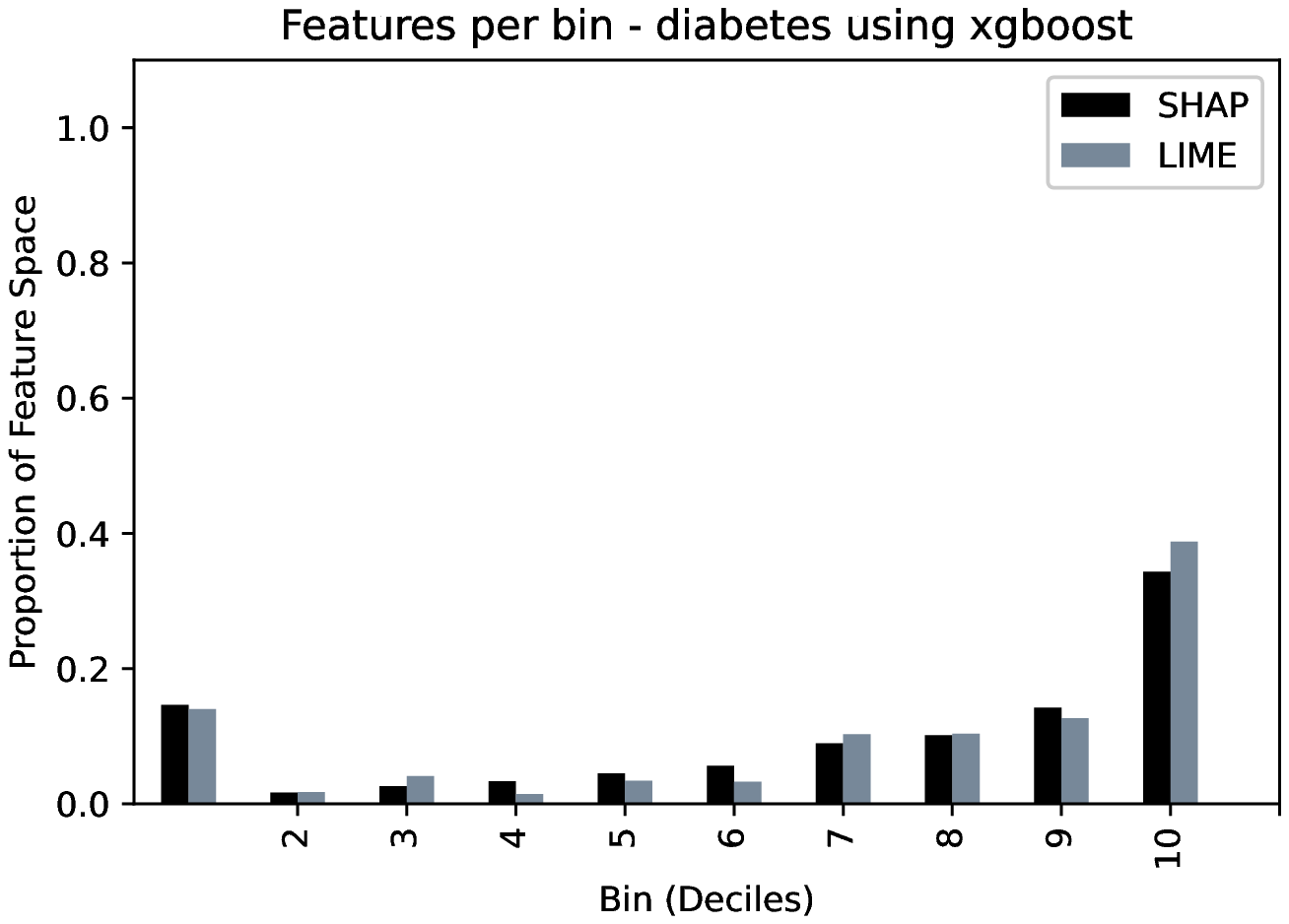}
    }
    
    \subfloat[Breast Cancer using Decision Tree]{
    \includegraphics[scale=0.4]{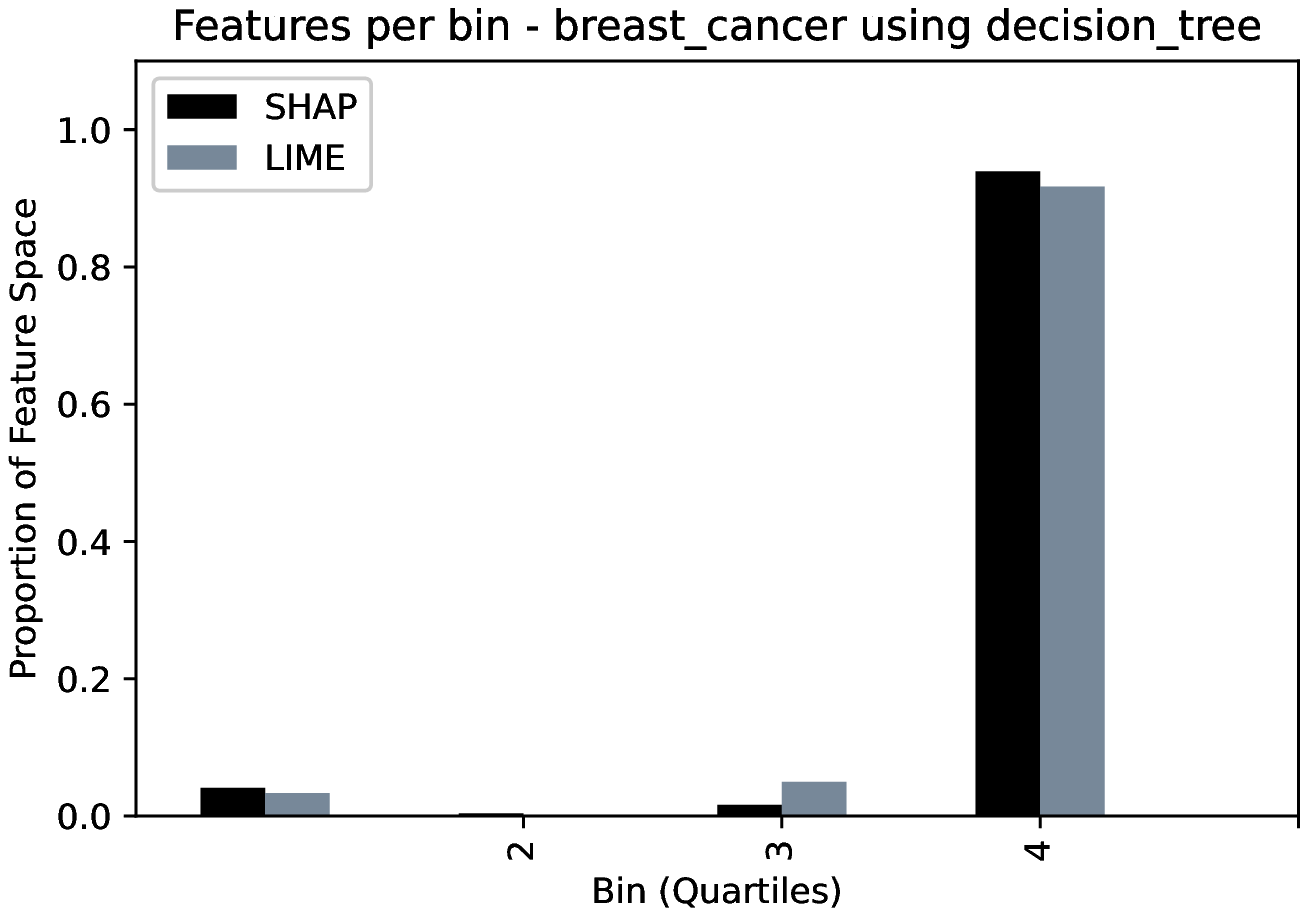}
    \includegraphics[scale=0.4]{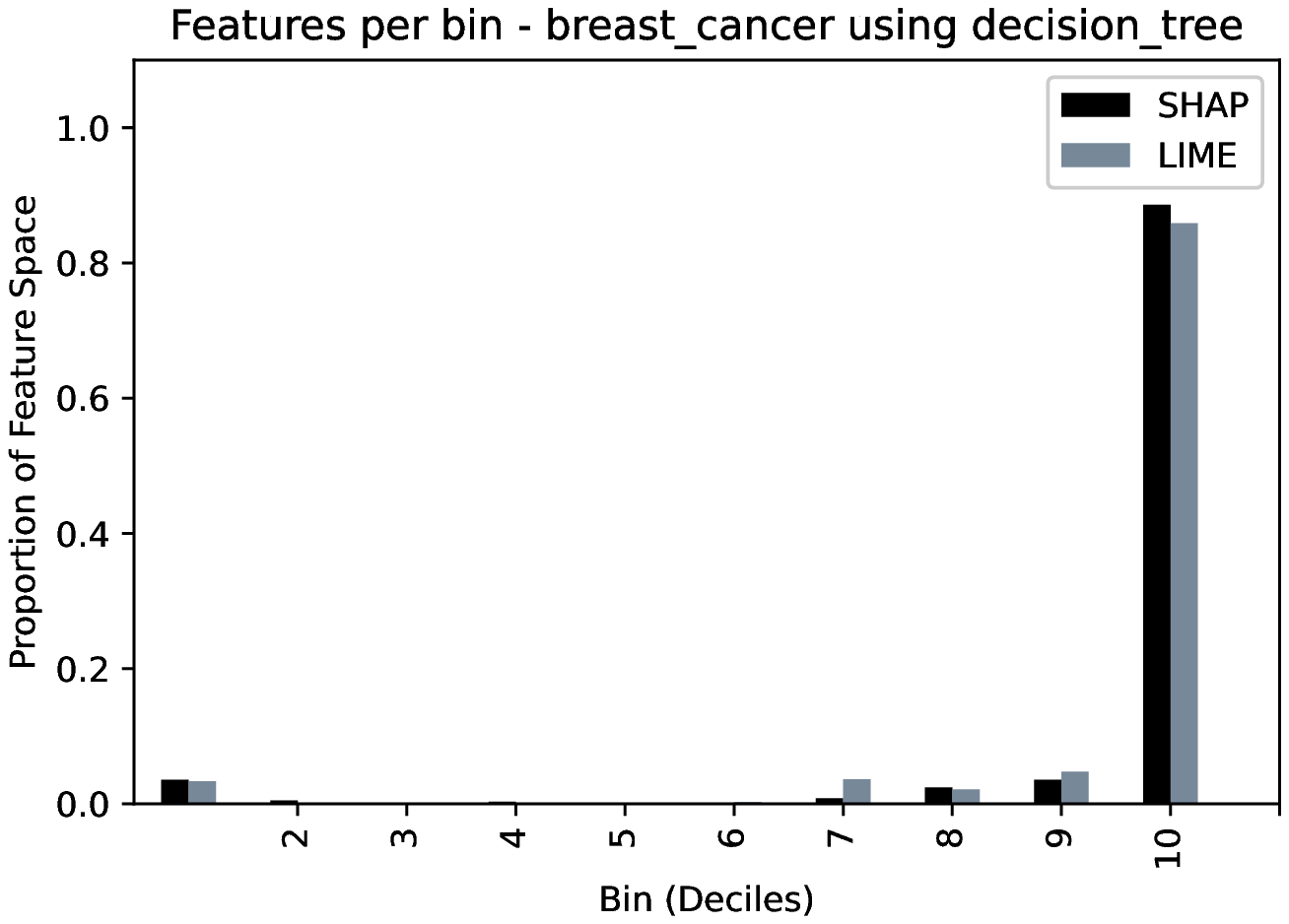}
    }\\
    \subfloat[Breast Cancer using XGBoost]{
    \includegraphics[scale=0.4]{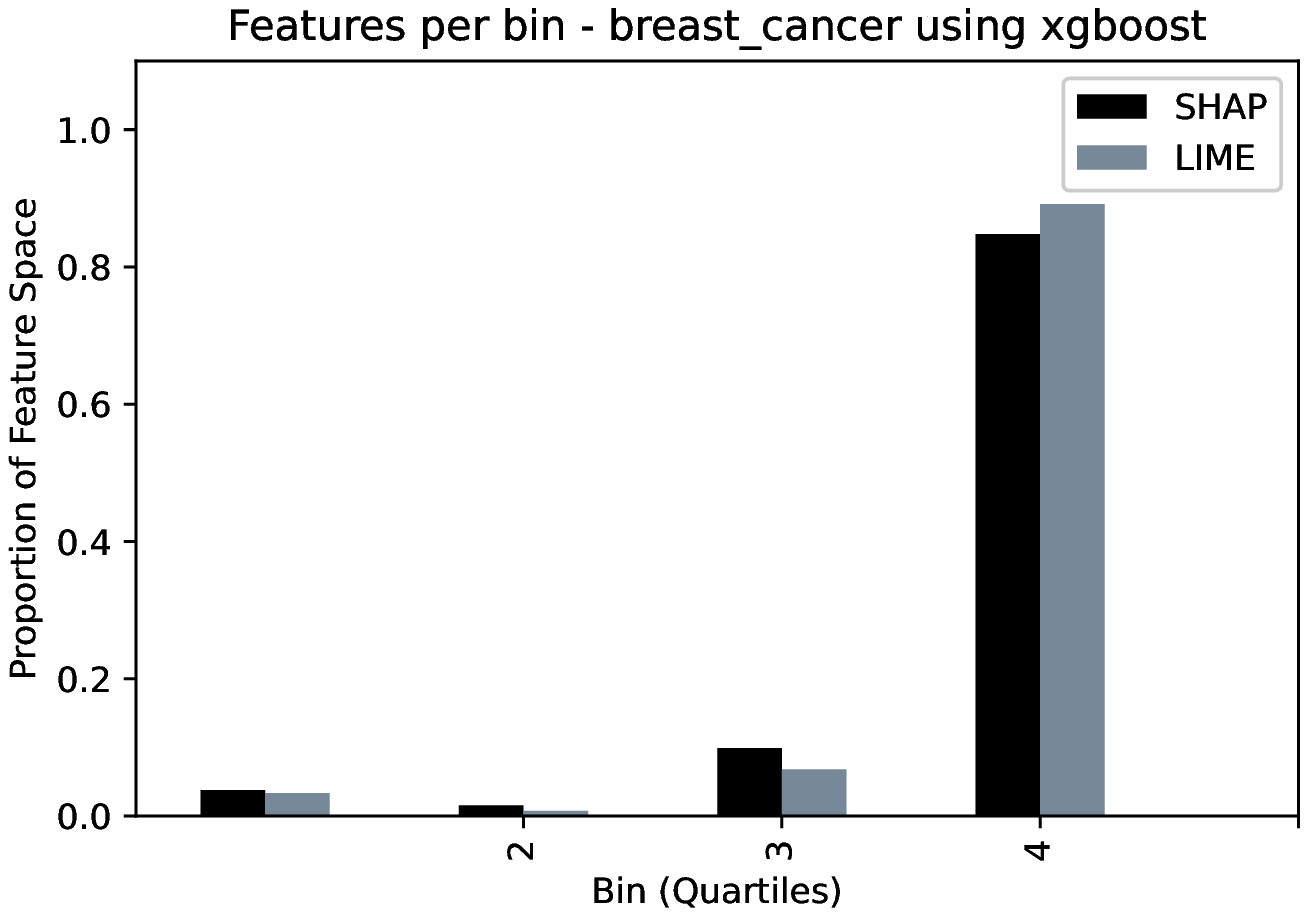}
    \includegraphics[scale=0.4]{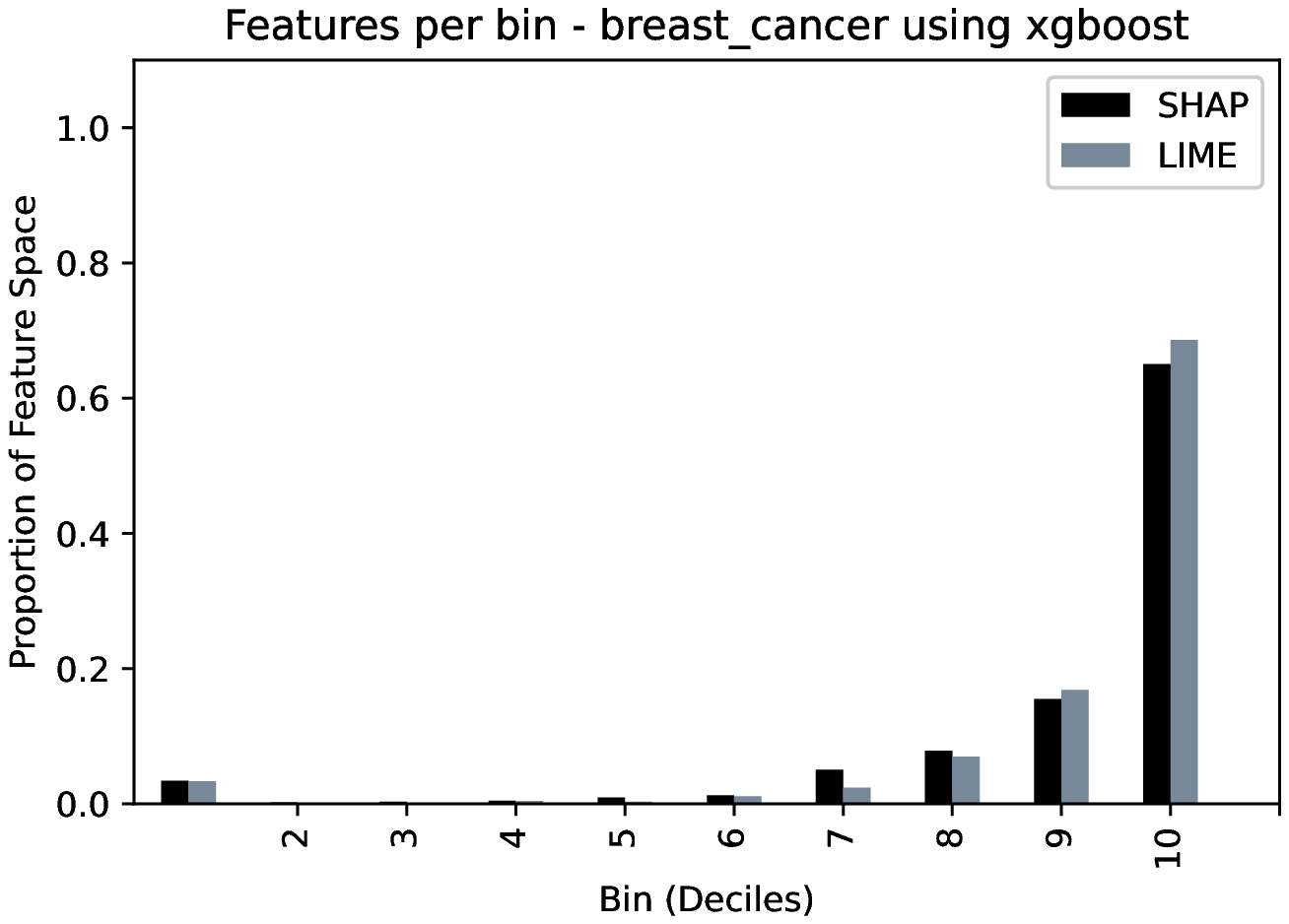}
    }
    \caption{Distribution of features by feature weight for the Diabetes and Student Results datasets}
    \label{fig:quartile_distributions}
\end{figure}

Predictive model complexity appears to have had little impact on the results in Phase 1. The Adult Income, Bike Sharing and House Price decision trees are the most complex out of six decision trees trained, but this does not seem to affect the recall and precision of LIME or SHAP. However, model complexity does seem to affect the parameters and results of the explanation-supporting and explanation-contrary methods. The results in Phase 2 showed that the number of optimal deciles were fewest for the Breast Cancer and Diabetes datasets, which also produced the simplest decision trees. It appears that, for LIME, the more complex the model becomes, the more distributed features are in the explanation. Interestingly, SHAP shows the opposite pattern. This is particularly apparent when comparing the two explainable methods' feature weight distribution for the diabetes and breast cancer decision trees (the simplest of all models created) against the more complicated XGBoost models for the same datasets models (see figure~\ref{fig:quartile_distributions}).

Model complexity seems to have played a larger role in the results for the classification datasets in phase 3, where most of the results are poorer than in phase. SHAP's explanation-contrary fidelity in particular is strongly affected. It appears that, with the more complex XGBoost model, the boundaries that can be identified through the explanation are weaker. In essence, feature values that produce dissimilar SHAP values to the original are still producing similar or same results when input into the predictive model. Also notably, LIME's explanation-contrary fidelity increases for the Adult Income dataset, for which it had the poorest explanation-contrary fidelity in the previous phase. This is possibly a result of the underlying predictive model's lack of complexity, given that this is the simplest of the three XGBoost classifiers.

This finding regarding predictive model complexity, to some degree, reflects the findings of ~\citet{Yalcin2021}, who found that explanation correctness is correlated with dataset complexity. This prior study also used tree-based predictive models, and it is likely that the complexity of the model is driven by the complexity of the dataset. The previous study also found that SHAP was generally more accurate than LIME, though our study suggests that both are more accurate in certain facets. Explanation-contrary fidelity, for example, was higher for LIME when using the more complex XGBoost classifiers, while SHAP had better explanation-supporting fidelity for those models. In many cases, there appears to be a trade-off, where the explainable method that was most effective for one type of fidelity was not the most effective in another in the same situation (for example, precision and recall in Phase 1).

It is also important to note that explanation-contrary fidelity was generally quite low. This is possibly a reflection of precision declining as more deciles of features are evaluated, though generally, it appears as though the boundaries implied by the explanation are not firm. This is important to note, as it can be easy to assume that the inverse of boundaries implied by explanations can be treated as a counterfactual explanation.


\subsection{Limitation and Future Work}\label{sec:future}
There are a few limitations associated with the described results. Firstly, while the three phases were applied to two different types of predictive models, both were tree-based models, and, as such, it becomes unclear whether these results will also hold true for other classes of models, such as deep learning models. In particular, other classes of models may yield more interesting results for regression problems, given the limitations of the decision-tree based models in this regard. Given that SHAP is also optimised on the basis of the underlying predictive model, including other classes of predictive models would also provide a more rounded evaluation of SHAP.

Future work should also consider other classes of explainable methods, such as probabilistic interpretable models~\citep{Moreira21dss}. In this work, we chose to evaluate LIME and SHAP as typical examples of local feature attribution methods. However, past studies with SHAP have shown that feature attribution explanations have little to no effect on user's trust of a black box model~\citep{Zhang2020} or, when attempting to determine the correctness of a machine-generated alert, little to no effect on task effectiveness or user's mental efficiency~\citep{Weerts2019}. Therefore, it's been suggested that comparative explanations or counterfactual explanations could be more useful than feature attribution methods in helping users understand and evaluate prediction algorithms~\citep{Weerts2019}. Given that a key goal in the field of XAI is to improve human understanding of algorithmic decision-making, it is also necessary to develop evaluation methods to evaluate such classes of explainable methods as a key step in ensuring explainable methods are truly of value, and to identify limits that exist in such methods. The three phase approach outlined in this work can be used to adapt the methods described here, or develop new approaches for evaluation.

Another limitation of this work was the use of heuristics to determine appropriate parameters to use when choosing the most relevant features and feature values to use in the explanation-supporting and explanation-contrary fidelity evaluations. Rather than relying on heuristics, future work aimed at developing optimisation functions to determine such parameters would be of use. In particular, more work is needed to understand the variations that were seen in the described results of phase 2, particularly between the regression and classification datasets. We theorise that this variation is due to model and data complexity, but future work exploring this in further detail could shed detail on choosing appropriate parameters for each situation, rather than relying on general heuristic as we have done in this work.

\section{Conclusion}~\label{sec:conclusion}
Given the increasing use of explainable AI methods to understand the decision-making of complex, opaque predictive models, it becomes necessary to understand how well an explainable method can interpret any given model. However, evaluation methods to achieve this remain an open question in the field of XAI, particularly for tabular data. In this work, we proposed the following: a three-phase approach to developing an evaluation method for XAI; an evaluation method to assess the explanations provided by feature attribution explainable methods for tabular data; and an evaluation of LIME and SHAP, two popular feature attribution methods, for models trained on six, well-known, open-source datasets.

As part of our study, we refined existing evaluation approaches into an evaluation method for assessing the fidelity of black box models trained on tabular data, and defined to measures of fidelity. Our evaluations of LIME and SHAP using this method revealed the internal mechanism of the underlying predictive model, the internal mechanism of the explainable method and model and data complexity all affected explanation fidelity. We also found that, given that explanation fidelity is so sensitive to context, there was no one explainable method that was clearly superior to the other, and that boundaries set by explanations were rarely firm. We also highlight the future work needed to further refine and extend the proposed approach and method.

\begin{acknowledgements}
Computational resources and services used in this work were provided by HPC and Research Support Group, Queensland University of Technology, Brisbane, Australia.
\end{acknowledgements}

\section{Declaration}
\subsection{Funding}
The lead author of this work is a PhD candidate who is currently receiving a stipend as part of the Australian Government's Research Training Program.
\subsection{Conflicts of interest/Competing interests}
The authors have no conflicts of interest or competing interests related to the submitted work.
\subsection{Availability of data and material}
All datasets used in this work are open source and available through the UCI Machine Learning Repository
\subsection{Code Availability} 
The scripts used to conduct the experiments described in this study are all available at \url{https://git.io/JZdVR}.
\subsection{Ethics approval} 
Ethics approval was not required for the submitted work.
\subsection{Consent to participate} 
There were no participants in the study described in this work.
\subsection{Consent for publication}
There were no participants in the study described in this work.

\bibliographystyle{spbasic}      
\bibliography{references}   

%
%

\end{document}